%% file: iclr2027_conference.tex
\documentclass{article}
\usepackage{iclr2027_conference,times}

\input{math_commands.tex}

\usepackage{amsmath,amssymb}
\usepackage{booktabs}
\usepackage{multirow}
\usepackage{float}
\usepackage{microtype}
\usepackage{graphicx}
\usepackage{tabularx}
\usepackage{enumitem}
\usepackage{xcolor}
\usepackage{tikz}
\usepackage{tabularx}
\usepackage{wrapfig}
\usepackage{float}
\usetikzlibrary{arrows.meta,fit,positioning,shapes.geometric}
\usepackage{hyperref}
\usepackage{url}
\hypersetup{hidelinks}

\definecolor{loopablue}{HTML}{246BCE}
\definecolor{loopborange}{HTML}{D97706}
\definecolor{guardgreen}{HTML}{18864B}
\definecolor{lightgray}{HTML}{F4F5F7}
\definecolor{draftred}{HTML}{B42318}

\newcommand{\method}{\textsc{EvoQuantLab}}

\newcommand{\status}[1]{\textsc{#1}}

\newcommand{\cP}{\mathcal{P}}

\newcommand{\cD}{\mathcal{D}}

\title{EvoQuantLab: Training-Free Dual-Loop Evolution
for Auditable Quantitative Research}

\author{%
{\bfseries Haolin Yang$^1$ \quad
Zhiqiang Qian$^1$ \quad
Chutian Chen$^1$ \quad
Jipeng Zhang$^2$ \quad
Zhenghao Zhu$^1$} \\
{\bfseries Tangjia Deng$^1$ \quad
Jian Xie$^3$ \quad
Shuaishuai Gong$^4$ \quad
Yike Guo$^1$\thanks{Corresponding authors.} \quad
Sirui Han$^1$\footnotemark[2]} \\
{\normalfont\small
$^1$The Hong Kong University of Science and Technology \quad
$^2$Microsoft \quad
$^3$Tsinghua University} \\
{\normalfont\small $^4$University of Macau}
}

\iclrfinalcopy
\begin{document}

\maketitle
% This is an arXiv preprint, not an ICLR-accepted conference paper.
\lhead{Preprint}

\input{sections/00_abstract}

\input{sections/01_introduction}

\input{sections/02_related_work}

\input{sections/03_method}
\input{sections/04_evaluation}

\input{sections/05_preliminary}

\input{sections/07_conclusion}

\subsubsection*{Reproducibility statement}
To support independent reproduction, we will release the source code, dependency
specifications, experiment configurations, and task adapters for the CSI~500,
S\&P~500, and hourly cryptocurrency benchmarks. The release will specify the exact
temporal splits, point-in-time universe construction, preprocessing and label
definitions, transaction-cost assumptions, random seeds, and evaluation procedures.
When source licenses prevent direct redistribution of market data, we will instead
provide data-acquisition and preparation scripts together with schemas and content
hashes needed to reconstruct and verify the experimental inputs. Each run records
versioned task and protocol manifests, candidate hashes, append-only trial and LLM-call
ledgers, cached predictions, and atomic checkpoints. We will also release anonymized
search trajectories and frozen OOS evaluation artifacts so that the reported results
and the separation between development and held-out evaluation can be audited.

\subsection*{AI use statement}
We used an AI assistant for grammar checking and translation during manuscript preparation. For grammar checking, we used the assistant to identify grammatical errors and suggest corrections to sentence structure, tense, and word usage. For translation, we used it to translate text, with attention to fluency, consistent technical terminology, and preservation of the original meaning.
\bibliography{iclr2027_conference}
\bibliographystyle{iclr2027_conference}

\appendix
\input{sections/appendix}

\end{document}

%% file: math_commands.tex
\usepackage{amsmath,amsfonts,bm}

\def\eqref#1{equation~\ref{#1}}
\def\1{\bm{1}}

\DeclareMathAlphabet{\mathsfit}{\encodingdefault}{\sfdefault}{m}{sl}
\SetMathAlphabet{\mathsfit}{bold}{\encodingdefault}{\sfdefault}{bx}{n}

%% file: sections/00_abstract.tex
\begin{abstract}
Automated alpha factor discovery is critical for modern quantitative research. While large language models (LLMs) have increasingly been integrated into quantitative workflows, existing agent-based approaches remain highly susceptible to severe backtest overfitting. To address this challenge, we propose \method{}, a training-free, dual-loop self-evolving framework for robust factor discovery. Instead of relying entirely on LLM generation, \method{} deploys an inner genetic-programming (GP) loop that dramatically improves token efficiency while evolving typed factor expressions under structural and diversity constraints. An outer diagnosis loop then systematically audits evidence across cycles to drive protocol-level improvements. Furthermore, our evaluation pipeline applies multi-metric validation to rigorously filter out spurious factors before deployment. Extensive out-of-sample experiments across both daily and hourly frequencies---covering equities and cryptocurrency---demonstrate the robustness of \method{}. Specifically, on daily CSI~500 and S\&P~500 tasks, \method{} achieves leading annualized returns of 18.34\% and 21.30\%, consistently outperforming existing agent baselines. On high-turnover hourly cryptocurrency data, \method{} delivers a 37.91\% annualized return while effectively avoiding execution collapse.
\end{abstract}

%% file: sections/01_introduction.tex
\section{Introduction}
\label{sec:introduction}
Automated alpha discovery has become an increasingly important direction in quantitative research~\citep{gong2026artificial,xiao2024tradingagents,lopez2026can}. Recent large language model (LLM) agents can formulate economic hypotheses, generate factor implementations, interpret backtest feedback, and iteratively refine candidate signals, enabling substantially larger research spaces than traditional manual workflows~\citep{rdagentquant2025,alphaagent2025,quantaalpha2026,factorminer2026}. Systems such as R\&D-Agent-Quant, AlphaAgent, QuantaAlpha, and FactorMiner demonstrate that LLMs can automate significant portions of factor and model discovery across heterogeneous financial data.

\begin{figure}[ht]
\centering
\includegraphics[width=\textwidth]{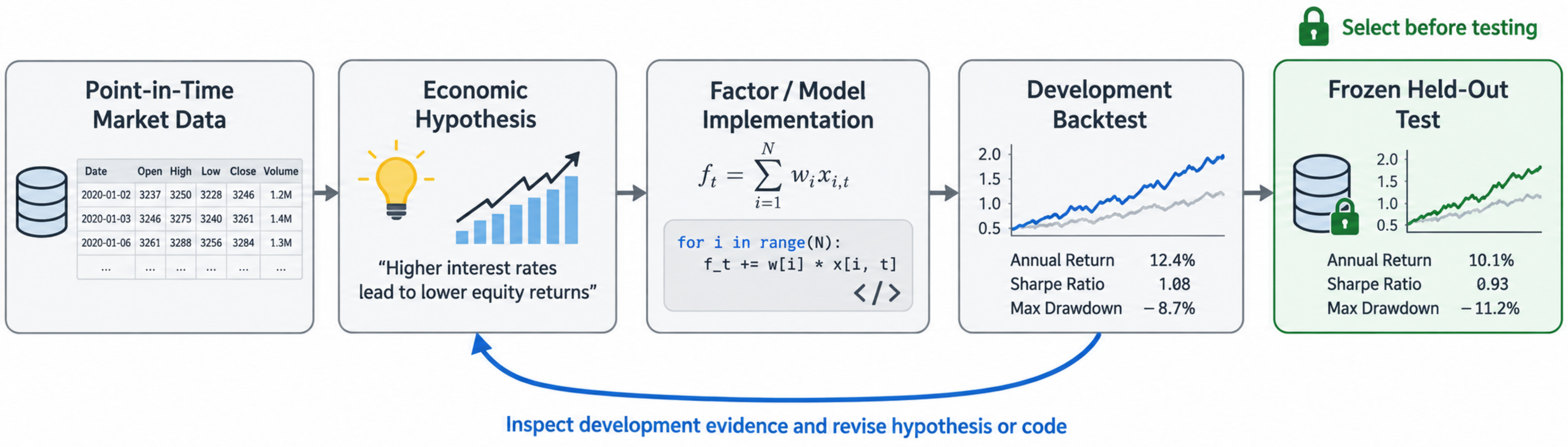}
\vspace{-0.6cm}
\caption{A conventional quantitative-research workflow. Iteration over development
backtests is necessary, but repeated inspection makes the complete search history part
of the statistical experiment. The held-out test must remain outside this feedback
path.}
\label{fig:traditional-workflow}
\vspace{-6pt}
\end{figure}
However, scaling automated research also amplifies a fundamental problem: \emph{backtest overfitting}. An agent may repeatedly generate and evaluate large numbers of factors, transformations, windows, and model variants, making it increasingly likely that apparently strong signals arise from noise or adaptive selection rather than persistent predictive structure~\citep{harvey2016crosssection,bailey2014deflated}. This issue is particularly severe for iterative LLM agents, since development results are repeatedly fed back into subsequent generations. Existing systems mainly focus on improving candidate generation, memory, reflection, or S\&Psearch~\citep{yang2024finrobotopensourceaiagent}, but stronger search alone can further increase the effective number of trials~\citep{kou2024automate}. Meanwhile, relying on LLM calls for every candidate is unnecessarily expensive when many factor variations are symbolic transformations that can be explored more efficiently~\citep{shi2025navigatingalphajunglellmpowered,RomeraParedes2023MathematicalDF}.

To address these challenges, we introduce \method{}, a training-free dual-loop self-evolving framework for robust alpha discovery. At the inner level, \method{} combines LLM-generated economic hypotheses with genetic programming (GP), which efficiently evolves typed factor expressions through selection, crossover, and mutation while maintaining structural and population diversity. At the outer level, the framework audits completed research cycles for evidence of overfitting, instability, excessive complexity, and selection bias, and uses these diagnoses to improve the research protocol for subsequent cycles. Importantly, protocol changes only affect future cycles rather than retroactively modifying completed evaluations. Together with a multi-metric validation pipeline that filters unstable, redundant, and economically inconsistent candidates, this design allows the search process itself to evolve while preserving disciplined out-of-sample evaluation.

We evaluate \method{} across daily and hourly quantitative prediction tasks covering equities and cryptocurrency. On CSI~500 and S\&P~500, \method{} achieves annualized returns of 18.34\% and 21.30\%, respectively, compared with 14.02\% and 15.23\% from the strongest baselines. On hourly cryptocurrency data, it achieves a 37.91\% annualized return despite the shorter horizon and substantially higher turnover. Beyond final performance, our experiments show that GP provides an efficient mechanism for large-scale factor exploration, while outer-cycle diagnosis improves robustness across successive research cycles.

Our contributions are:
\begin{itemize}[leftmargin=*,itemsep=2pt,topsep=3pt]
    \item We introduce \method{}, a training-free dual-loop framework that jointly evolves factor expressions and the quantitative research process to address adaptive backtest overfitting.
    \item We combine LLM-guided hypothesis generation with diversity-preserving genetic programming, enabling efficient exploration of large symbolic factor spaces with substantially reduced reliance on repeated LLM inference.
    \item We develop a diagnosis-driven outer evolution mechanism and rigorous multi-metric validation pipeline, and demonstrate strong out-of-sample performance across daily and hourly equity and cryptocurrency tasks.
\end{itemize}

%% file: sections/02_related_work.tex
\section{Related Work}
\vspace{-0.2cm}
\label{sec:related}

\paragraph{Automated alpha factor discovery.}

Formulaic alpha discovery seeks interpretable expressions that predict future asset returns. Classical approaches rely on human-designed signals or search over symbolic expressions, with genetic programming (GP) providing a natural mechanism for evolving executable factor trees through crossover and mutation \citep{koza1992gp,kronberger2026guidingmultiobjectivegeneticprogramming}. Evolutionary search has been applied to trading-rule discovery \citep{neely1997gp,allen1999genetic}, while AutoAlpha extends this paradigm with hierarchical evolutionary search for formulaic alphas \citep{autoalpha2020}.
Learning-based approaches instead optimize the generation process directly; for example, deep symbolic regression frameworks~\citep{petersen2019deep}and AlphaGen~\citep{alphagen2023} formulate formulaic alpha discovery as reinforcement learning. Despite their effectiveness, these methods typically operate within
a predefined search space whose operators and search procedure remain fixed. \method retains an explicit GP search engine but allows validated research knowledge
and the search vocabulary to evolve across cycles.

\paragraph{LLM agents for quantitative research.}

Recent work uses large language models to automate broader parts of quantitative research, connecting hypothesis generation, factor implementation, empirical evaluation, and iterative refinement\citep{wang-etal-2024-llmfactor,yang2026mars-sql}. R\&D-Agent(Q) coordinates research and development agents for joint factor--model optimization \citep{rdagentquant2025}; AlphaAgent combines LLM-driven factor generation with regularization against redundancy, complexity, and
alpha decay \citep{alphaagent2025}; and systems such as QuantaAlpha and FactorMiner further explore iterative factor discovery with evolutionary search and accumulated research experience \citep{quantaalpha2026,factorminer2026,li2023tradinggptmultiagentlayeredmemory}. These systems demonstrate the potential of LLMs as autonomous quantitative researchers, but also expose the model repeatedly to noisy backtest feedback in open-ended trial-and-error loops, where context noise readily misguides planning decisions\citep{lu2024ai,liang2026lost}. In contrast, \method{} separates semantic research evolution from candidate-level optimization: the LLM updates reusable research knowledge, while a deterministic, budget-accounted GP engine performs factor search.

\paragraph{Self-evolving agents.}
Beyond finance, self-evolving agents aim to improve their own reasoning procedures, workflows, architectures, or procedural skills through accumulated execution feedback and experiential memory
\citep{shinn2023reflexionlanguageagentsverbal,pmlr-v306-he26ar,zhao2024expelllmagentsexperiential,zheng2026skillprox}. Pioneering efforts search over executable programs and algorithmic heuristics using LLMs in an evolutionary loop\citep{romeraparedes2024mathematical,wang2023voyageropenendedembodiedagent,he2026clintutor}. Other methods automate agent designs and workflows via iterative feedback\citep{hu2025automateddesignagenticsystems,zhang2025aflowautomatingagenticworkflow,zhang2025exesql}, or recombine modular components like planning and memory\citep{shang2025agentsquareautomaticllmagent,zhang2026memcore}. This line of work shows that adaptation can occur at the system level rather than solely through model parameters, extending from agent workflows to broader research automation that co-evolves
skills, memory, and policies~\citep{xu2026nanoresearch}. \method{} brings this perspective to quantitative research, but evolves
research knowledge and symbolic search across cycles under a frozen evidence protocol, preventing held-out outcomes from influencing subsequent evolution.

%% file: sections/03_method.tex
\section{EvoQuantLab}
\label{sec:method}

\begin{figure}[t]
\centering
\includegraphics[width=\textwidth]{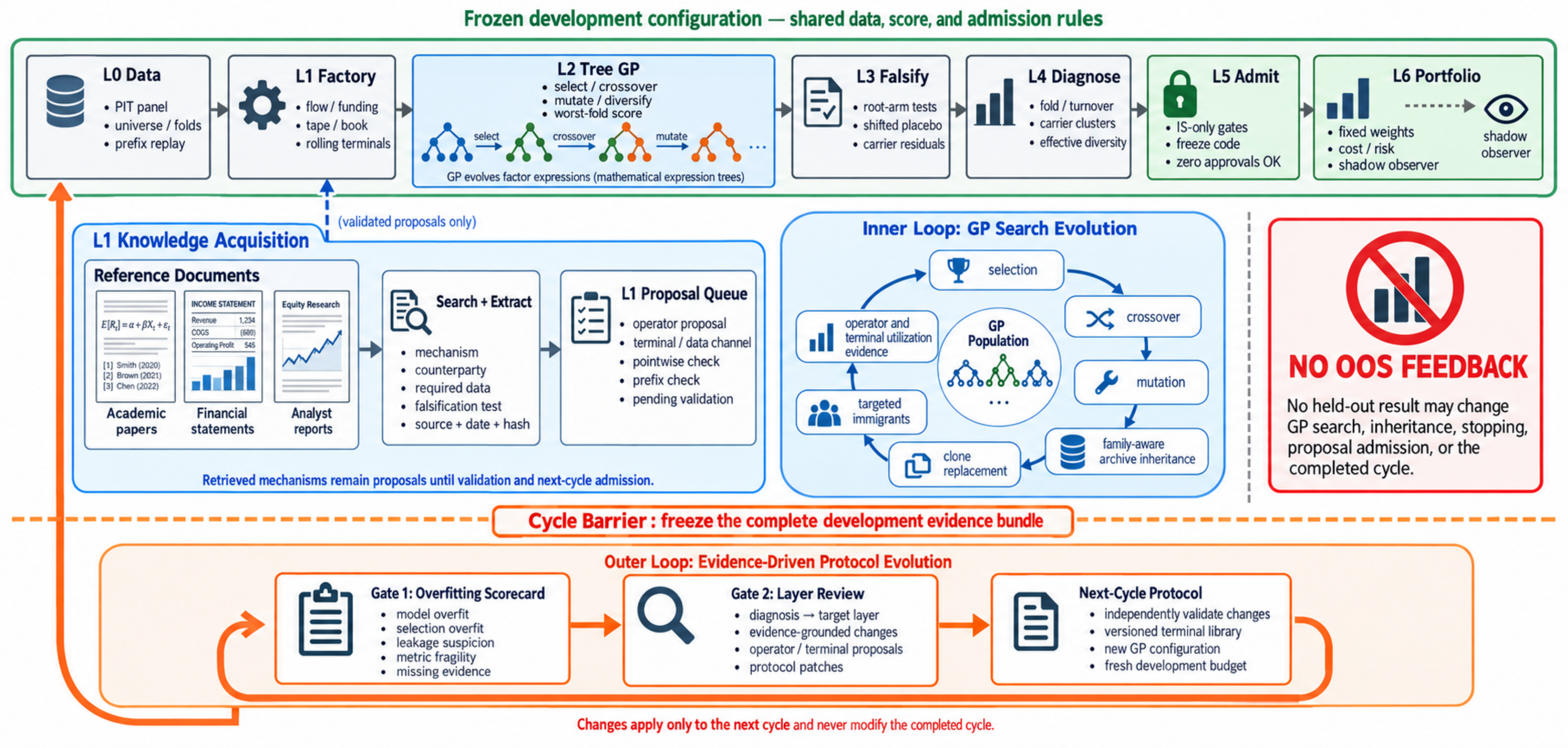}
\vspace{-0.5cm}
\caption{Overview of \method{}. A frozen L0--L6 evidence pipeline evaluates every candidate under a fixed data and evaluation contract. The fast inner loop evolves factor expressions and reallocates search capacity using development-only evidence, while the slow outer loop diagnoses overfitting after each completed research cycle and proposes a versioned protocol for the next cycle. Information from held-out evaluation never writes back into the current cycle.
}
\label{fig:overview}
\vspace{-7pt}
\end{figure}
\vspace{-0.2cm}
Automated quantitative discovery faces two intertwined challenges: efficiently exploring vast symbolic spaces, and preventing adaptive backtest overfitting caused by repeated experimentation.
\method{} addresses this by decoupling exploration across two distinct timescales: an \emph{inner fast loop} evolving factor expressions via genetic programming (GP) within a frozen protocol, and an \emph{outer slow loop} auditing research evidence across cycles to update protocol rules. 

\vspace{-5pt}
\subsection{Formal Setup \& Immutable Task Contract}
\label{sec:problem-setup}

Let $\cD = (\cD^{\mathrm{train}}, \cD^{\mathrm{dev}}, \cD^{\mathrm{oos}})$ be chronologically ordered market data. 
A research cycle $c$ is governed by a frozen protocol $\Pi_c = \langle \mathcal{S}_c, \mathcal{T}_c, \mathcal{O}_c, \mathcal{M}_c, \mathcal{B}_c \rangle$, where $\mathcal{S}_c$ specifies the task and data contract, $\mathcal{T}_c$ denotes the active terminal set, $\mathcal{O}_c$ the typed operator set, $\mathcal{M}_c$ the evaluation rules, and $\mathcal{B}_c$ the search and trial budgets.
To eliminate lookahead bias, $\Pi_c$ enforces point-in-time availability and \emph{prefix replay invariance}: for any $\tau < \tau'$, features computed on $\cD_{1:\tau}$ must agree with those computed on $\cD_{1:\tau'}$ over the shared prefix.
Within cycle $c$, \method{} maintains a population of typed factor trees $\cP_c = \{g_1,\dots,g_K\}$, where $K$ denotes the population size and each candidate $g \in \mathcal{G}(\mathcal{T}_c,\mathcal{O}_c)$ belongs to the valid expression space induced by the active terminal and operator sets. When evaluated on an asset--time observation, $g$ produces a scalar factor score.
Candidates are evaluated across $F$ development folds via a robust worst-fold objective penalized by structural tree complexity:
\begin{equation}
f_{\mathrm{dev}}(g; \Pi_c) = \min_{1 \le k \le F} \operatorname{Metric}_k(g) - \lambda \operatorname{Complexity}(g),
\label{eq:objective}
\end{equation}
where $\operatorname{Metric}_k(g)$ denotes the task-specific utility on development fold $k$, defined such that larger values indicate better performance; $\operatorname{Complexity}(g)$ is a pre-registered scalar complexity penalty derived from tree size and depth, and $\lambda>0$ controls its strength.
Equation~\ref{eq:objective} operates strictly on $\cD^{\mathrm{dev}}$; the holdout split $\cD^{\mathrm{oos}}$ is physically inaccessible during search. Training data $\cD^{\mathrm{train}}$ are used only for task-specific parameter estimation when required (e.g., fitting normalization or baselines), whereas candidate selection and protocol evolution are driven entirely by development evidence.

\vspace{-5pt}
\subsection{The Frozen L0--L6 Evidence Pipeline}
\label{sec:evidence-spine}

Candidates are sequentially validated through a cycle-frozen seven-layer pipeline (Fig.~\ref{fig:overview}), advancing only upon passing preceding gates.
The pipeline begins at \textbf{L0 (Data)}, which assembles canonical panels, tradable universes, and purged fold masks under strict prefix invariance. 
Next, \textbf{L1 (Factory)} instantiates typed, economically grounded terminals spanning order flow, funding, microstructure, and momentum, while \textbf{L2 (Tree GP)} searches typed symbolic expression trees guided by the worst-fold fitness objective in Equation~\ref{eq:objective}.
% (\S\ref{sec:inner-evolution})

Surviving expressions undergo economic falsification at \textbf{L3 (Falsify)} through only-arm/drop-arm ablations, sign-flip and participant-swap tests, and cross-sectional residualization against pre-registered factor carriers. A candidate survives only if its hypothesized core mechanism remains directionally consistent and the full expression provides reproducible incremental value.
At each timestamp, we cross-sectionally residualize candidate scores against a pre-registered carrier matrix and re-evaluate the residual signal across development folds. Candidates whose predictive power vanishes after residualization are rejected as redundant with known factor families.

Subsequently, \textbf{L4 (Stress Battery)} stress-tests signal stability across regime splits (market trend $\times$ volatility), execution delay curves, asset clusters, and matched-budget block permutations. 
Candidates that clear all development hurdles reach \textbf{L5 (Admit)}, where their implementation, terminal manifest, data version, and metric configuration are frozen and recorded by content hashes before OOS access. 
Finally, \textbf{L6 (Portfolio)} maps admitted signals into trading strategies under fixed volatility targeting, risk constraints, and transaction costs.

\vspace{-5pt}
\subsection{Loop A: Inner GP Search \& Knowledge Ingestion}
\label{sec:inner-evolution}

\textbf{L1 Knowledge Ingestion.}
Instead of expanding features blindly, L1 extracts economic mechanisms from academic papers, financial disclosures, and analyst reports into structured \emph{Reference Cards}. 
These proposals enter an isolated \emph{L1 Proposal Queue}. A candidate primitive must clear both a pointwise reference check and a prefix causality test before becoming eligible for admission across cycle boundaries. 
Each active terminal $j \in \mathcal{T}_c$ maintains a development evidence tuple $e_j = (u_j, s_j, r_j, d_j)$, tracking utilization $u_j$, candidate survival $s_j$, residual contribution $r_j$, and cross-fold stability $d_j$. Primitives with weak standalone power but sustained interaction gains are retained as \emph{interaction-only} to preserve diversity.

\textbf{Tree-Based Factor Evolution.}
Within cycle $c$, population $\cP_c$ evolves across generations via tournament selection. 
Offspring are generated via typed subtree crossover and point mutation over valid grammar nodes. 
Syntactic duplicates and expressions with pairwise correlation $\rho > 0.95$ are replaced with targeted immigrants from underrepresented families. Elites are preserved in a family-stratified archive instead of a monolithic top-$K$ list, guaranteeing structural diversity within the trial budget.

\vspace{-4pt}
\subsection{Loop B: Diagnosis-Driven Protocol Evolution}
\label{sec:outer-evolution}

The outer loop activates only after an entire cycle crosses the \textbf{Cycle Barrier} (Fig.~\ref{fig:overview}). Unlike the inner GP loop, Loop~B does not modify candidates within the current cycle. Instead, it audits accumulated training and development evidence and proposes bounded protocol changes that take effect only in the next cycle. Importantly, both diagnosis and protocol evolution are performed without access to the held-out test set.

\textbf{Gate 1: ML Overfitting Scorecard. }
Gate~1 audits execution traces for evidence of adaptive overfitting, including
train-to-dev degradation, cross-seed variance, tree complexity, concentration, label sensitivity, and search multiplicity.
To control for selection multiplicity, let $m_t^{\mathrm{real}} = f_{\mathrm{dev}}(g_t; \Pi_c)$ denote the selection score of the $t$-th evaluated candidate under protocol $\Pi_c$, and define $m^{\mathrm{real}}(T) = \max_{1 \le t \le T} m_t^{\mathrm{real}}$ as the best score achieved across $T$ real search trials.
We run $B_0$ independent null pseudo-searches with the same trial budget under a
pre-registered block-permutation null, and record
$m_b^{\mathrm{null}}(T) = \max_{1 \le t \le T} m_{b,t}^{\mathrm{null}}$.
\begin{equation}
\hat{p}_{\mathrm{sel}}(T)
=
\frac{
1 + \sum_{b=1}^{B_0}
\mathbf{1}\!\left[
m_b^{\mathrm{null}}(T)
\ge
m^{\mathrm{real}}(T)
\right]
}{
B_0 + 1
}.
\label{eq:selection-p}
\end{equation}
Equation~\ref{eq:selection-p} evaluates whether the search procedure itself yields alpha beyond a budget-matched random search over noise. Gate~1 outputs a diagnostic status $\mathcal{V}_{\mathrm{diag}} \in \{ \mathtt{Pass},\allowbreak \mathtt{Overfit\text{-}Model},\allowbreak \mathtt{Overfit\text{-}Selection},\allowbreak \mathtt{Suspect\text{-}Leakage},\allowbreak \mathtt{Metric\text{-}Fragile},\allowbreak \mathtt{Insufficient} \}$.

\textbf{Gate 2: Routed Review \& Child Protocol Synthesis. }

Gate~2 uses $\mathcal{V}_{\mathrm{diag}}$ to route the review toward the protocol components most likely responsible for the observed failure mode. For example, suspected leakage triggers an audit of L0 data availability and split construction; model overfitting routes to complexity control; selection overfitting routes to the search space and trial budget; and metric fragility routes to robustness analysis of the evaluation criteria. The reviewer then proposes a bounded patch set $\mathcal{A}_c$, whose changes are applied only to the next protocol version:
\begin{equation}
\Pi_{c+1} = \operatorname{Apply}(\Pi_c, \mathcal{A}_c), \qquad h_{c+1} = \operatorname{Hash}(\Pi_{c+1}).
\label{eq:protocol-update}
\end{equation}
Completed results from cycle $c$ remain permanently associated with $\Pi_c$; the patch set never retroactively modifies candidate scores, selection decisions, or evaluation rules from the completed cycle.
Appendix~\ref{app:case-studies} provides case studies illustrating how \method{} redirects unproductive search, rejects deceptively strong candidates, and validates model-suggested knowledge before admission.
% \paragraph{OOS Governance: No Write-Back.}
% Held-out evaluation against $\cD^{\mathrm{oos}}$ is governed by a read-only \emph{shadow observer} (Fig.~\ref{fig:overview}). Once L5 freezes a candidate digest, a single-read token is executed and irrevocably logged. Crucially, held-out observations are strictly forbidden from writing back: they cannot rescue rejected candidates, tune parameters, or alter protocol $\Pi_c$. OOS lessons only inform future child protocols evaluated on fresh holdout splits, preserving statistical isolation across cycles.

%% file: sections/04_evaluation.tex
\vspace{-5pt}
\section{Main Experiments}
\label{sec:evaluation}

We evaluate whether dual evolution improves out-of-sample performance under a reproducible evaluation protocol. We benchmark \method{} against four representative agent-based alpha discovery baselines: \textbf{AlphaAgent}~\citep{alphaagent2025}, which regularizes LLM factor exploration against redundancy and signal decay; \textbf{QuantaAlpha}~\citep{quantaalpha2026}, an evolutionary framework optimizing formulaic alphas through LLM-driven operators; \textbf{FactorMiner}~\citep{factorminer2026}, a self-evolving agent guided by accumulated skill and experience memory; and \textbf{RD-Agent-Quant}~\citep{rdagentquant2025}, a multi-agent framework coordinating data-centric factor mining and model optimization. For a fair comparison, all methods execute their native workflows using the identical foundation model (Claude Opus~5) while sharing the exact same point-in-time data, temporal splits, transaction-cost assumptions, and evaluation suite. The out-of-sample (OOS) data remain strictly frozen and unobserved during the factor discovery process, accessed solely during final validation.
\vspace{-4pt}
\subsection{Tasks and metrics}
\vspace{-4pt}
We evaluate all methods on three quantitative prediction tasks spanning diverse markets, asset classes, and temporal resolutions, as summarized in Table~\ref{tab:tasks}. The daily-bar benchmarks cover CSI~500 (China equities) and S\&P~500 (U.S. equities) for next-day cross-sectional return prediction. To examine generalization under higher turnover and noise, we evaluate on an hourly cryptocurrency benchmark for next-hour return prediction. 

\begin{table}[ht]
\caption{Summary of the three evaluation tasks.}
\label{tab:tasks}
\centering

\resizebox{0.95\textwidth}{!}{%
\begin{tabular}{@{}lclll@{}}
\toprule
Dataset & Frequency & Market & Train / Dev Period & Prediction target \\
\midrule
CSI~500
& Daily
& China equities
& 2025-03-31–2026-04-30
& Next-day cross-sectional return \\

S\&P~500
& Daily
& U.S. equities
& 2025-03-31–2026-04-30
& Next-day cross-sectional return \\

Crypto
& Hourly
& Digital assets
& 2025-04-01–2026-03-31
& Next-hour cross-sectional return \\
\bottomrule
\end{tabular}%
}
\vspace{-5pt}
\end{table}

We assess performance along two dimensions: factor predictive ability and realized portfolio quality. Predictive power is captured by the information coefficient (IC, cross-sectional Pearson correlation) and its temporal stability (ICIR). Portfolio performance is evaluated via an execution-aware long-short strategy with transaction costs, measured by annualized return (AR), cumulative return (Ret.), information ratio (IR), and maximum drawdown (MDD). Full trading and execution details are deferred to Appendix \ref{app:eval_details}.
\vspace{-4pt}
\subsection{Daily and hourly results}
\begin{figure*}[ht]
    \centering
    \includegraphics[width=\textwidth]{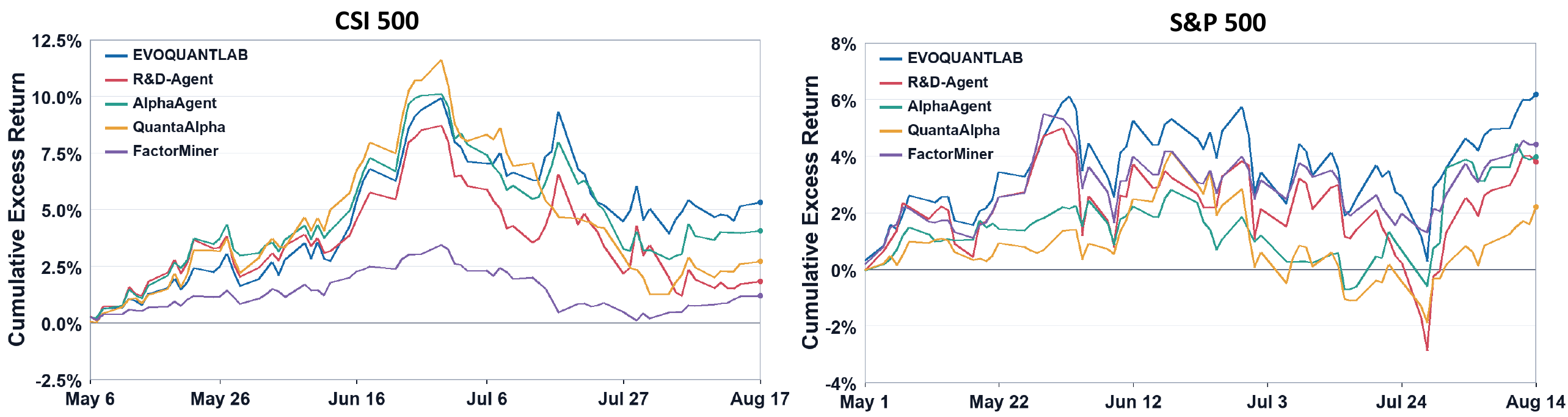}
    \vspace{-0.5cm}
    \caption{Cumulative excess returns of different approaches on CSI 500 and S\&P 500.}
    \label{fig:daily-return-curves}
\end{figure*}

\vspace{-4pt}
Table~\ref{tab:daily-results} presents the daily-bar results on CSI~500 and S\&P~500. On both equity markets, \method{} consistently outperforms baseline agent frameworks across predictive and risk-adjusted metrics. Specifically, on CSI~500, \method{} achieves the highest IC (0.0281), ICIR (0.1211), IR (1.613), and cumulative return (5.31\%), with a competitive MDD (-4.78\%). On S\&P~500, \method{} delivers top-tier performance with an IC of 0.0233, an IR of 1.614, and an annualized return of 21.30\%, while maintaining a well-controlled drawdown (-5.82\%). Figure~\ref{fig:daily-return-curves} depicts the corresponding cumulative excess-return trajectories.

\vspace{-10pt}
\begin{table}[ht]
\caption{Main experimental results on CSI 500 and S\&P 500. \textbf{Bold} indicates the best result.}
\label{tab:daily-results}
\centering
\scriptsize
\setlength{\tabcolsep}{2.2pt}

\resizebox{\textwidth}{!}{%
\begin{tabular}{@{}l*{12}{c}@{}}
\toprule
& \multicolumn{6}{c}{\textbf{CSI 500} (2026-05-06--2026-08-17)}
& \multicolumn{6}{c}{\textbf{S\&P 500} (2026-05-01--2026-08-14)} \\
\cmidrule(lr){2-7}\cmidrule(l){8-13}
Method
& IC $\uparrow$ & ICIR $\uparrow$ & IR $\uparrow$ & AR $\uparrow$ & Ret. $\uparrow$ & MDD $\downarrow$
& IC $\uparrow$ & ICIR $\uparrow$ & IR $\uparrow$ & AR $\uparrow$ & Ret. $\uparrow$ & MDD $\downarrow$ \\
\midrule
AlphaAgent
& 0.0177 & 0.1072 & 1.340 & 14.02\% & 4.06\% & -7.09\%
& 0.0070 & 0.0573 & 1.490 & 13.67\% & 3.96\% & \textbf{-3.48\%} \\
QuantaAlpha
& 0.0139 & 0.0893 & 0.939 & 9.41\% & 2.73\% & -9.88\%
& 0.0095 & 0.0663 & 0.796 & 7.59\% & 2.20\% & -5.92\% \\
FactorMiner
& 0.0187 & 0.0746 & 1.047 & 4.11\% & 1.19\% & \textbf{-3.26\%}
& 0.0173 & 0.0954 & 1.582 & 15.23\% & 4.41\% & -4.18\% \\
RD-Agent
& 0.0080 & 0.0438 & 0.546 & 6.31\% & 1.83\% & -7.36\%
& 0.0136 & 0.0580 & 0.902 & 13.07\% & 3.79\% & -7.76\% \\
\method
& \textbf{0.0281} & \textbf{0.1211} & \textbf{1.613} & \textbf{18.34\%} & \textbf{5.31\%} & -4.78\%
& \textbf{0.0233} & \textbf{0.1029} & \textbf{1.614} & \textbf{21.30\%} & \textbf{6.17\%} & -5.82\% \\
\bottomrule
\end{tabular}%
}
\end{table}
\vspace{-0.5cm}

\begin{figure*}[ht]
\centering
\begin{minipage}[t]{0.48\textwidth}
    \vspace{-9pt}
    \centering
    \scriptsize
    \setlength{\tabcolsep}{3.0pt}
    \renewcommand{\arraystretch}{1.25} 
    \begin{tabular}{@{}lcccccc@{}}
\toprule
& \multicolumn{6}{c}{\textbf{Crypto} (2026-04-01--2026-06-09)} \\
\cmidrule(l){2-7}
Method
& IC $\uparrow$
& ICIR $\uparrow$
& IR $\uparrow$
& AR $\uparrow$
& Ret. $\uparrow$
& MDD $\downarrow$ \\
\midrule
AlphaAgent
& 0.0366 & 0.1278 & -2.385 & -27.50\% & -5.20\% & -12.40\% \\
QuantaAlpha
& 0.0320 & 0.1472 & -1.665 & -10.66\% & -2.02\% & -4.55\% \\
FactorMiner
& 0.0086 & 0.0482 & 0.952  & 9.53\%   & 1.80\%   & -3.48\% \\
RD-Agent
& \textbf{0.0444} & \textbf{0.1807} & -1.870 & -16.15\% & -3.05\% & -8.29\% \\
\method{}
& 0.0089 & 0.0299 & \textbf{3.709} & \textbf{37.91\%} & \textbf{7.17\%} & \textbf{-3.23\%} \\
\bottomrule
\end{tabular}
\end{minipage}
\hfill
\begin{minipage}[t]{0.46\textwidth}
    \vspace{-11pt}
    \centering
    \includegraphics[width=\linewidth]{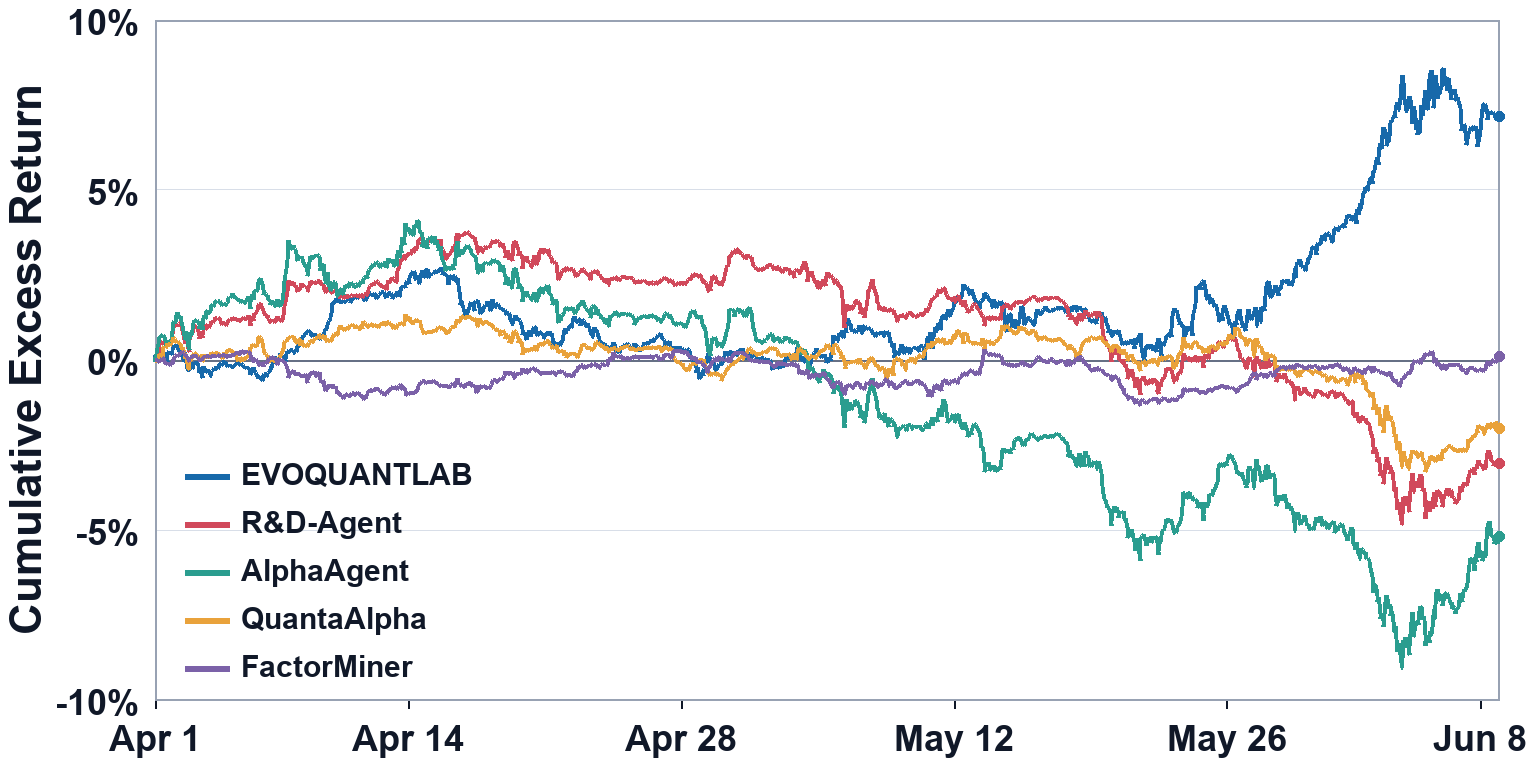}
\end{minipage}
\caption{Performance comparison on Crypto. Left: predictive and portfolio
metrics of each method. Right: cumulative
excess-return trajectories over the OOS period.}
\label{fig:crypto-results}
\end{figure*}

The higher-frequency Crypto benchmark reveals a stark divergence between predictive correlation and execution profitability. As shown in Figure~\ref{fig:crypto-results}, several agent baselines generate high predictive correlations on paper but collapse during execution: RD-Agent achieves the highest IC (0.0444) and ICIR (0.1807), yet yields a negative IR (-1.870) and substantial drawdowns (-8.29\%). In contrast, \method{} successfully converts discovered signals into robust trading returns, achieving the highest IR (3.709), cumulative return (7.17\%), and the lowest drawdown (-3.23\%). This highlights \method{}'s ability to discover executable, noise-resilient alphas in high-turnover regimes.

\vspace{-5pt}
\subsection{Dev-to-OOS Generalization and Overfitting Audit}
\label{sec:overfitting-audit}
\vspace{-5pt}
A central challenge in automated alpha discovery is backtest overfitting, where repeated model inspection and candidate selection produce deceptive in-sample performance that collapses out-of-sample. To directly evaluate whether discovered signals exhibit true out-of-sample persistence, we perform a pre-registered transfer audit. For each method, we identify the single best-performing factor strictly using development data ($\text{Dev IR}$), freeze its implementation, and evaluate the \emph{identical} factor on the held-out OOS period across all three markets without any re-tuning.

\begin{wraptable}{r}{0.51\textwidth}
\centering
\small
\setlength{\tabcolsep}{4pt}
\vspace{-10pt}
\caption{Out-of-sample IR of the single development-selected factor. Means reflect equal weighting across markets.}
\label{tab:dev_oos_transfer}
\resizebox{\linewidth}{!}{%
\begin{tabular}{@{}lcccc@{}}
\toprule
\textbf{Method} & \textbf{Crypto} & \textbf{CSI 500} & \textbf{S\&P 500} & \textbf{Mean} \\
\midrule
\textbf{\method{}}      & \textbf{4.055} & \textbf{1.041} & \textbf{2.466} & \textbf{2.521} \\
RD-Agent-Quant         & -0.229          & 1.002           & -0.082          & 0.230          \\
AlphaAgent             & -2.385          & 0.798           & -0.490          & -0.692         \\
QuantaAlpha            & -4.999          & -0.079          & 0.819           & -1.420         \\
FactorMiner$^\dagger$  & 0.952           & 0.547           & 1.382           & 0.960          \\
\bottomrule
\end{tabular}%
}
\vspace{2pt}

\parbox{\linewidth}{\scriptsize$^\dagger$FactorMiner retains only 1 candidate per market, bypassing candidate selection.}
\vspace{-10pt}
\end{wraptable}

As reported in Table~\ref{tab:dev_oos_transfer}, \method{} achieves remarkably consistent positive out-of-sample performance across all three markets, delivering an average OOS IR of $2.521$ and preserving positive alpha ($+\to+$ transition) in 100\% of the tasks. In contrast, existing agent-based baselines exhibit substantial performance degradation: RD-Agent, AlphaAgent, and QuantaAlpha suffer severe IR collapses and sign reversals on both equity and high-frequency crypto benchmarks, reflecting that iterative LLM feedback inadvertently optimizes against development noise. While FactorMiner maintains positive returns, it retains only a single candidate factor per market ($M=1$), completely bypassing the multiple-testing selection problem. 

Furthermore, as detailed in the comprehensive audit in Appendix~\ref{app:overfitting_diagnostics}, \method{} achieves an average positive-sign survival rate of $85.5\%$ across all retained candidates (reaching $95.5\%$ on CSI~500). This confirms that our outer-loop diagnostic gating and frozen evidence pipeline effectively suppress selection bias and prevent spurious signals from surviving to deployment.

%% file: sections/05_preliminary.tex
\vspace{-6pt}
\section{Ablation Studies}
\label{sec:ablation}

% We organize the ablations around the current GP-centered design: evolution across
% cycles, GP updates within a cycle, and knowledge acquisition at L1. Unless stated
% otherwise, all variants use identical
% data, folds, terminal manifests, finalist gates, random seeds, and candidate-evaluation
% budgets. We report development and frozen evaluation metrics separately and aggregate
% results over at least five independent runs.
\vspace{-6pt}
\subsection{Component Ablation Study}
\label{sec:ablation-components}

\begin{wraptable}{r}{0.7\textwidth}
\centering
\small
\setlength{\tabcolsep}{4.5pt}
\vspace{-0.6cm}
\caption{Component ablation on the dual-loop framework. Metrics report out-of-sample performance of the top L5-approved factor chosen via a predeclared selection rule. Values marked with ``--'' indicate that no sampled candidates passed the validation filters to reach L5 approval.}
\label{tab:component-ablation}
\begin{tabular}{lccccc}
\toprule
Variant & IC $\uparrow$ & ICIR $\uparrow$ & IR $\uparrow$ & AR $\uparrow$ & Ret. $\uparrow$ \\
\midrule
Full \method{}             & \textbf{0.0089} & \textbf{0.0299} & \textbf{3.709} & \textbf{37.91\%} & \textbf{7.17\%} \\
Cycle~0 only               & 0.0067          & 0.0343          & 1.570          & 19.93\%          & 3.77\%          \\
Cycle~0 w/o GP updates     & --              & --              & --             & --               & --               \\
No Loop B & 0.0055 & 0.0281  & 0.752 &  6.58\% & 1.54\% \\
No L1 search               & 0.0039          & 0.0120          & 2.331          & 27.44\%          & 5.19\%           \\
No diversity controller    & -0.0029          & -0.0156          & 1.712          & 10.51\%          & 2.58\%           \\

\bottomrule
\end{tabular}
\end{wraptable}
Table~\ref{tab:component-ablation} reports the performance of key architectural components under a strictly matched candidate-evaluation budget.
\emph{Cycle~0 only} terminates after the initial iteration, isolating the cumulative benefits of cross-cycle inheritance and evidence-guided refinement. 
\emph{Cycle~0 without GP updates} ablates iterative genetic search (selection, crossover, and mutation) in favor of uniform random sampling; this yields no qualified factors that satisfy our multi-stage pipeline filters, underscoring that directed evolutionary optimization is essential to factor viability.
\emph{No Loop~B} retains multi-cycle execution but removes the outer diagnosis loop, preventing protocol-level feedback and evidence-driven search adjustments across cycles.
\emph{No L1 search} fixes the terminal and operator spaces to their initial states, quantifying the marginal gain of dynamic operator retrieval across cycles. 
Finally, \emph{No diversity controller} removes family-stratified inheritance, clone deduplication, and targeted immigrant injection, directly measuring their impact against premature population collapse.

Across all variants, all metrics reflect frozen out-of-sample evaluations. These results highlight two key insights: while directed GP updates are strictly necessary to generate valid, filter-passing alphas, dynamic L1 expansion, Loop~B outer diagnosis, and family diversity control are critical drivers for maximizing downstream predictive and risk-adjusted alpha quality.

\vspace{-5pt}
\subsection{Evolution across Cycles}
\label{sec:ablation-cycle-trajectory}

\begin{figure}[t]
\centering
\includegraphics[width=\textwidth]{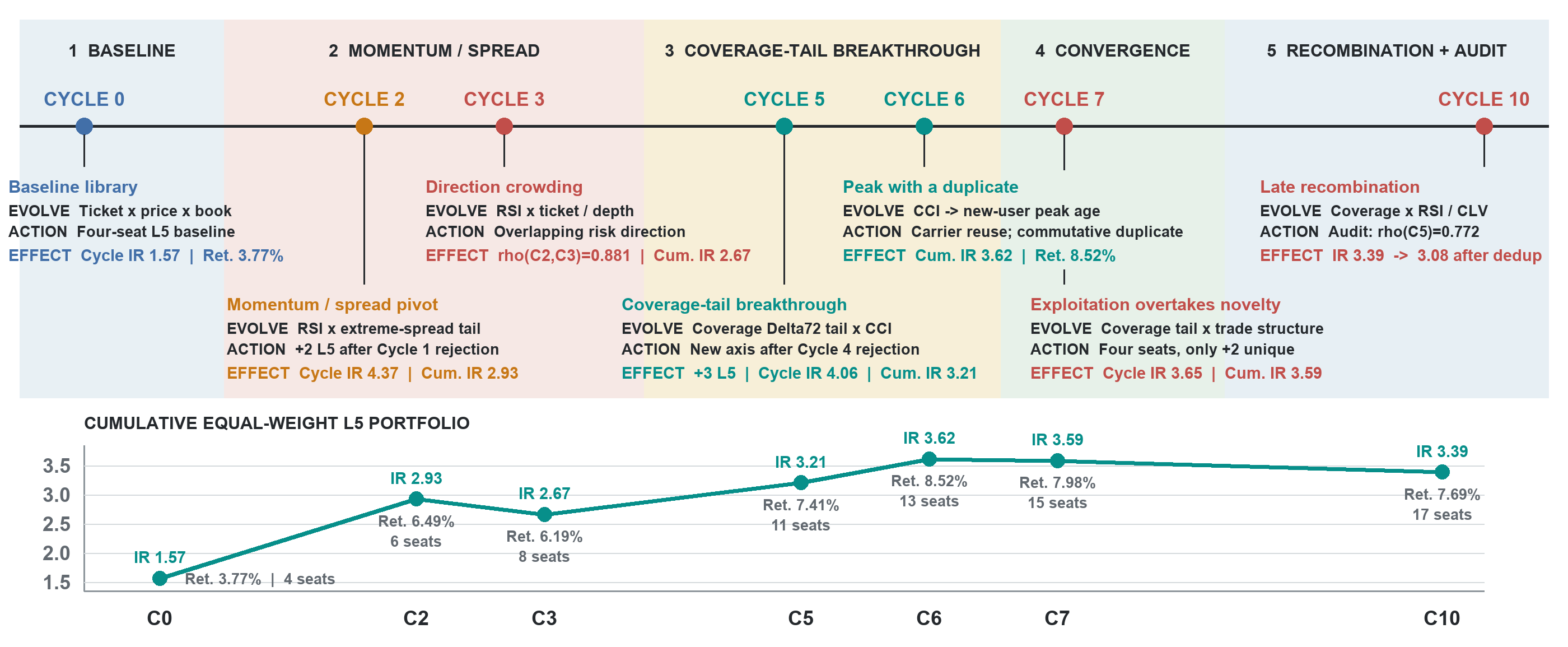}
\vspace{-18pt}
\caption{Retrospective trajectory of discovered factor families and the cumulative L5 portfolio across outer cycles on Crypto. Crucially, all OOS IR and Return values shown are computed strictly post-hoc for retrospective analysis; no OOS signal is exposed to the search agent.}
\label{fig:cycle-evolution}
\vspace{-10pt}
\end{figure}

We further trace how the system evolves across outer cycles rather than comparing only the initial and final states. Figure~\ref{fig:cycle-evolution} tracks a representative Crypto run, depicting key search events and the cumulative equal-weight portfolio of L5-approved factors to contrast genuine search-space expansion with redundant family refinement.

The trajectory exhibits a clear transition from exploration to exploitation. Initialized with four factors at Cycle~0 (IR 1.57), early evolution targets momentum- and spread-based structures, elevating the cumulative IR to 2.93 by Cycle~2. While Cycle~3 finds extra candidates, their highly overlapping risk profile ($\rho=0.881$) yields negligible gain, showing that factor quantity alone cannot bolster the portfolio.

A larger improvement occurs in the middle of the run when evolution expands into a
previously underexplored factor family. After the rejected Cycle~4 proposals, Cycle~5
introduces coverage-tail signals combined with CCI and contributes three new
L5-approved factors, increasing the cumulative IR from 2.67 to 3.21. Continued
refinement of this family reaches the peak portfolio IR of 3.62 at Cycle~6, together
with an arithmetic return of 8.52\%. These transitions suggest that the main benefit
of iterative evolution comes from discovering new predictive axes rather than merely
recombining candidates within an already saturated region of the search space.

Later cycles show diminishing returns and increasing redundancy. By Cycle~7,
exploitation begins to dominate novelty: four portfolio seats are proposed but only
two correspond to unique additions, while the cumulative IR slightly decreases from
3.62 to 3.59. The final recombination stage exhibits a similar pattern. Although
Cycle~10 continues to construct candidates from coverage, RSI, and CLV signals, the
audit identifies substantial overlap with earlier discoveries; deduplication reduces
the corresponding IR from 3.39 to 3.08. Overall, the trajectory shows that the
evolution process is most effective while it continues to open new factor families,
whereas later cycles increasingly revisit correlated structures and approach a
performance plateau.

\subsection{Evolution Efficiency Analysis}
\label{sec:ablation-efficiency}
We evaluate the execution and resource efficiency of \method{} alongside four representative agent baselines: R\&D-Agent-Quant, AlphaAgent, QuantaAlpha, and FactorMiner. All methods are evaluated under identical task settings, environment constraints, and hardware allocations. Table~\ref{tab:efficiency} details the end-to-end computational and financial footprint across wall time, responsiveness, LLM token and API invocation budgets, and factor discovery throughput.
\vspace{-5pt}
\begin{table}[ht]
\caption{Efficiency and resource consumption comparison across factor discovery methods. API costs are computed using a standardized pricing schedule frozen across all runs.}
\label{tab:efficiency}
\centering
\small
\setlength{\tabcolsep}{3.2pt}
\resizebox{\textwidth}{!}{%
\begin{tabular}{@{}lrrrrrrr@{}}
\toprule
Method & Wall time & Time to first valid & Input tokens & Output tokens
& LLM calls & API cost & L5 factors / hour \\
\midrule
R\&D-Agent-Quant & 18m39s & 1m47s & 91745 & 18007 & 26 & 0.909 & 45 \\
AlphaAgent       & 25m39s & 5m39s & 149870 & 67721 & 23 & 2.442 & 9.36 \\
QuantaAlpha      & 2h4m & 2m29s & 767897 & 455379 & 94 & 15.224 & 10.2 \\
FactorMiner      & 24m36s & 50s & 102771 & 11040 & 21 & 0.790 & 156 \\
\method{}        & 22h08m & 2h45m & 520780 & 256000 & 50 & 9.000 & 0.95 \\
\bottomrule
\end{tabular}%
}
\vspace{-10pt}
\end{table}
\vspace{-8pt}
\paragraph{Time and Search Depth Trade-offs.}
As reported in Table~\ref{tab:efficiency}, lightweight baselines such as FactorMiner and R\&D-Agent-Quant terminate rapidly (under 25 minutes) and quickly produce initial candidate factors (under 2 minutes), yielding higher nominal throughput (up to 156 L5 factors per hour). However, this rapid completion stems from constrained search horizons and relatively shallow validation cycles. In contrast, \method{} sustains an extended, deep-horizon exploration trajectory spanning 22h08m, with the first valid candidate materializing at 2h45m. This reflects the thoroughness of \method{}'s iterative refinement and multi-stage evaluation pipeline, prioritizing deliberate, high-conviction factor evolution over superficial high-volume sampling.

\paragraph{Token and Financial Budget Economy.}
Despite running substantially longer than iterative competitors, \method{} maintains a moderate computational and financial footprint. Specifically, \method{} incurs only 50 LLM calls with 520,780 input and 256,000 output tokens, translating to an API cost of \$9.000. In comparison, QuantaAlpha incurs over 767K input tokens, 455K output tokens, and 94 API calls over its 2-hour run, escalating costs to \$15.224 without engaging in multi-round hypothesis evolution of comparable depth. This demonstrates that \method{} achieves its exhaustive search regime primarily through rigorous numerical and symbolic validation rather than wasteful LLM token churn.

% \begin{figure}[H]
% \centering
% \fbox{\begin{minipage}[c][31mm][c]{0.94\linewidth}
% \centering
% \textit{Planned efficiency curves}\\[2mm]
% Incumbent quality versus wall-clock time.\\
% Incumbent quality versus cumulative tokens and API cost.\\
% Valid and L5-approved candidates versus wall-clock time.
% \end{minipage}}
% \caption{Anytime efficiency of \method{} and the four agent baselines under matched
% hardware, concurrency, data, and candidate-evaluation budgets. Curves are aggregated
% over independent runs with uncertainty bands.}
% \label{fig:efficiency}
% \end{figure}

\subsection{Backbone Model Analysis}
\label{sec:ablation-backbone}

\begin{wrapfigure}{r}{0.4\textwidth}
\centering
\vspace{-12pt}
\includegraphics[width=\linewidth]{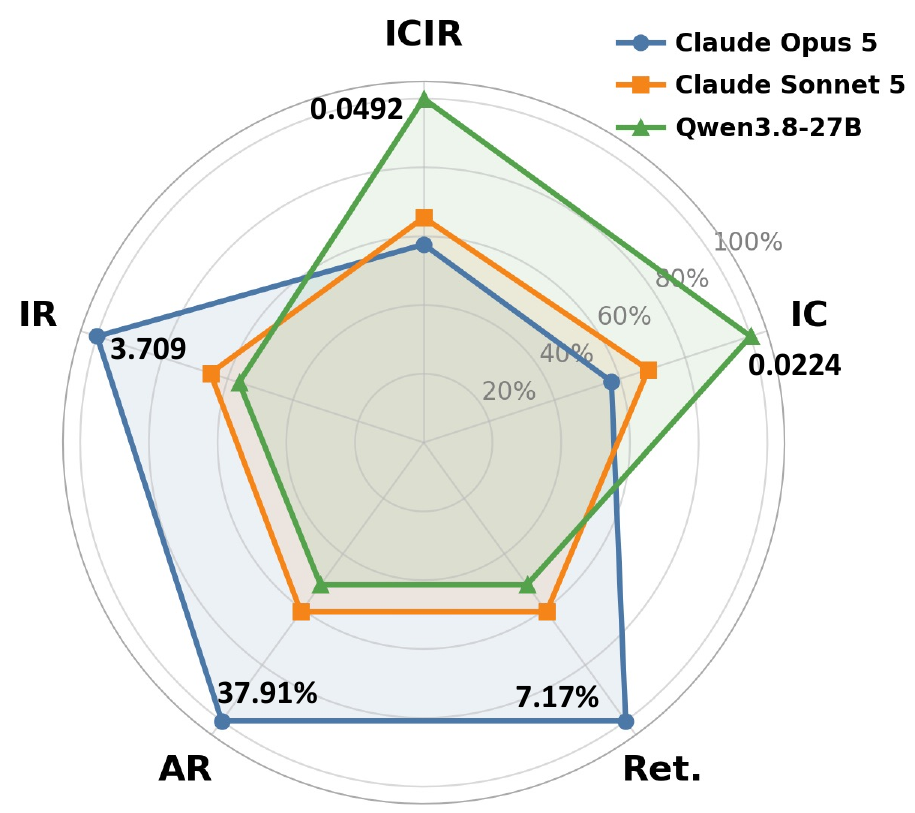}
\caption{Comparison of \method{} using different models
(11 cycles each). Axes are normalized, with outer points indicating better performance.}
\label{fig:backbone-radar}
\vspace{-10pt}
\end{wrapfigure}

\paragraph{Backbone Comparisons.}
To assess sensitivity to the underlying backbone, we benchmark Claude Opus5, Claude Sonnet5, and the open-weight Qwen3.8-27B (Fig.\ref{fig:backbone-radar}). Claude Opus5 achieves the strongest portfolio-level performance, particularly in risk-adjusted returns and proposal admission, reflecting better structural adherence and generalization. Claude Sonnet~5 delivers competitive intermediate performance across dimensions. Qwen3.8-27B, despite achieving the highest standalone IC, exhibits a lower admission rate and a larger gap between predictive strength and downstream portfolio performance. This pattern suggests that smaller backbones are more prone to overfitting during factor discovery, producing signals that appear strong under localized predictive metrics but generalize less reliably through the full validation pipeline. Nevertheless, its strong raw IC highlights the potential of open-weight models for cost-efficient, locally deployed alpha discovery.

%% file: sections/07_conclusion.tex
\section{Conclusion}
\label{sec:conclusion}

In this paper, we introduced \textsc{EvoQuantLab}, a training-free, dual-loop self-evolving framework designed to tackle the critical challenge of backtest overfitting and trial multiplicity in automated quantitative research. By decoupling token-efficient symbolic exploration in an inner genetic-programming loop from evidence-driven protocol evolution in an outer diagnostic loop, \textsc{EvoQuantLab} establishes an auditable, causally principled quantitative research workflow under strict immutable task contracts and one-way OOS governance.

Comprehensive evaluations across daily equities (CSI 500 and S\&P 500) and high-frequency digital assets demonstrate that \textsc{EvoQuantLab} significantly outperforms representative agent-based baselines, consistently translating predictive correlation into superior realized, risk-adjusted returns. The ablation studies and longitudinal cycle analyses confirm that directed genetic evolution, dynamic knowledge acquisition, and multi-metric falsification are indispensable for preventing population collapse and weeding out spurious signals. Remarkably, the framework achieves these robustness gains while maintaining a modest LLM token and financial footprint. We believe this paradigm of disciplined, multi-timescale self-evolution offers a robust, reproducible foundation for next-generation autonomous scientific and quantitative discovery.

%% file: sections/appendix.tex
\section{Dual-Loop Execution Procedure}
\label{app:algorithm}

For completeness, the outer-cycle procedure is given below. All state transitions are
deterministic except candidate and patch proposals made by the frozen API model.

\begin{enumerate}[leftmargin=*,itemsep=2pt]
    \item Load protocol $\Pi_c$ and verify its content hash. Instantiate the task adapter
    from the immutable task specification.
    \item Run the unchanged L0--L6 GP search engine on train/dev data. Freeze the
    complete cycle artifact bundle. Abort outer review if the inner run is incomplete.
    \item Construct per-terminal evidence from use, survival, ablation, residual, fold,
    and redundancy artifacts. Loop A emits lifecycle and budget decisions. Optional
    reference search emits proposal cards only.
    \item Construct a \texttt{CandidateAuditInput} for every finalist. Gate~1 emits an
    \texttt{MLOverfitReport} and a priority-ordered list of target layers.
    \item Gate~2 reviews layer evidence in that order. It emits typed local patches with
    evidence paths, expected benefit, risks, and a next-cycle validation plan.
    \item Reject patches that touch an immutable field, lack evidence, lack a regression
    test, or attempt to change the completed cycle. Apply admitted patches to create
    $\Pi_{c+1}$ and compute its hash.
    \item Atomically write the Loop A review, Gate~1 reports, Gate~2 review, next
    manifest, and outer checkpoint. Resume only from the last atomic checkpoint.
\end{enumerate}

\section{Gate 1 Evidence Schema}
\label{app:gate1}

\begin{table}[h]
\caption{Minimum Gate~1 evidence. Thresholds are calibrated per task before OOS access.}
\centering
\small
\begin{tabularx}{\textwidth}{@{}lXX@{}}
\toprule
Dimension & Required fields & Typical red flag \\
\midrule
Generalization & train, inner validation, outer development in one metric & sign reversal or small outer/train ratio \\
Fold/seed stability & worst fold, dispersion, same-sign rate, rank consistency & winner identity changes with seed \\
Complexity & feature/parameter count, simple-model score, one-SE loss & simpler model is statistically tied \\
Selection & total and family trials, matched-budget permutation winners & real winner is ordinary under null search \\
Concentration & date/entity contribution and leave-group-out results & a few dates or assets explain the score \\
Placebo & shifted or block-permuted labels & placebo remains comparable to real score \\
Label/metric fragility & adjacent horizons and pre-registered metrics & isolated horizon peak or ranking reversal \\
Trading mapping & prediction bins, gross/net PnL, turnover and cost curve & Sharpe lacks monotonic predictive support \\
\bottomrule
\end{tabularx}
\end{table}

The reference implementation uses conservative defaults only for contract tests. The
protocol manifest records every calibrated value, its calibration data, and the commit
preceding all OOS reads.

\section{Evaluation Details}
\label{app:eval_details}

\subsection{Metrics}
We adopt two classes of metrics to evaluate quantitative performance: factor-level predictive metrics and strategy-level portfolio returns.

\paragraph{Information Coefficient (IC).}
IC measures the cross-sectional correlation between the model's predicted scores and the subsequent realized asset returns. Following standard practice in quantitative finance, it is defined as:
\begin{equation}
\text{IC}_t = \frac{(\hat{y}_t - \mathbb{E}[\hat{y}_t])^\top (y_t - \mathbb{E}[y_t])}{\sigma(\hat{y}_t) \cdot \sigma(y_t)}
\end{equation}
where $\hat{y}_t$ and $y_t$ denote the predicted factor scores and realized forward returns across all assets at time step $t$, respectively; $\mathbb{E}[\cdot]$ denotes the cross-sectional mean, and $\sigma(\cdot)$ is the standard deviation. The reported IC is the time-series average across all evaluation steps $T$:
\begin{equation}
\text{IC} = \frac{1}{T} \sum_{t=1}^T \text{IC}_t
\end{equation}

\paragraph{Information Coefficient Information Ratio (ICIR).}
ICIR quantifies the temporal stability and consistency of the predictive signal across evaluation periods:
\begin{equation}
\text{ICIR} = \frac{\text{mean}(\text{IC}_t)}{\text{std}(\text{IC}_t)}
\end{equation}
A higher ICIR indicates more robust and reliable predictive quality over time.

\paragraph{Cumulative Return (Ret.).}
Cumulative return measures the aggregate percentage gain of the strategy portfolio across the entire evaluation horizon:
\begin{equation}
\text{Ret.} = \prod_{t=1}^{T} (1 + r_t) - 1
\end{equation}
where $r_t$ is the portfolio return at step $t$ after accounting for transaction costs.

\paragraph{Annualized Return (AR).}
AR reflects the annualized compound growth rate of the portfolio:
\begin{equation}
\text{AR} = (1 + \text{Ret.})^{\frac{N}{T}} - 1 = \left( \prod_{t=1}^T (1 + r_t) \right)^{\frac{N}{T}} - 1
\end{equation}
where $N$ is the number of periods in a year. We set $N = 252$ for daily equity tasks (CSI~500 and S\&P~500) and $N = 8760$ ($24 \times 365$) for the hourly cryptocurrency task.

\paragraph{Information Ratio (IR).}
IR assesses the risk-adjusted excess return of the strategy relative to its return volatility:
\begin{equation}
\text{IR} = \frac{\text{mean}(r_t - r_b)}{\text{std}(r_t - r_b)} \times \sqrt{N}
\end{equation}
where $r_b$ is the benchmark return. Under our dollar-neutral long-short portfolio setup where excess return is measured with $r_b = 0$, IR corresponds to the annualized Sharpe ratio of the zero-investment portfolio.

\paragraph{Maximum Drawdown (MDD).}
MDD measures the peak-to-trough decline of the portfolio value over the evaluation horizon, capturing tail downside risk:
\begin{equation}
\text{MDD} = - \max_{t \in [1, T]} \left( \frac{\max_{\tau \in [1, t]} P_\tau - P_t}{\max_{\tau \in [1, t]} P_\tau} \right)
\end{equation}
where $P_t = \prod_{\tau=1}^t (1 + r_\tau)$ denotes the cumulative portfolio net value at time $t$. We report MDD with a negative sign to reflect the percentage loss.
\subsection{Trading Strategy}
\label{app:trading_strategy}

We implement execution-aware portfolio simulations tailored to the characteristics of equities and digital assets.

\paragraph{Daily Equities (CSI~500 and S\&P~500).}
For equity markets, all methods utilize a daily-bar LightGBM paradigm based on pre-screened feature pools. The trading pipeline is simulated as follows:
\begin{itemize}
    \item \textbf{Pre-OOS Feature Selection}: Candidate factors undergo strict pre-screening using only development data (train and validation sets). We remove exact duplicates, discard factors with negative validation IC or IR, and greedily prune redundant factors based on the Pearson correlation of their daily portfolio weights (candidates with an absolute correlation $\ge 0.70$ against previously accepted factors are excluded). This retains 9 factors for CSI~500 and 20 factors for S\&P~500.
    \item \textbf{Model Training and Target}: Features are rank-normalized and mean-centered cross-sectionally each day. The LightGBM model is trained to predict the next-day cross-sectional return ranking, with hyperparameters and early stopping (patience of 50) determined strictly on the validation set without OOS lookahead (CSI~500: max depth 4, 15 leaves, learning rate 0.02; S\&P~500: max depth 5, 31 leaves, learning rate 0.03).
    \item \textbf{Portfolio Construction}: At the close of trading day $t$, the model generates score $s_{i,t}$ for stock $i$. The dollar-neutral portfolio weight $w_{i,t}$ is constructed as:
    \begin{equation}
    w_{i,t} = \frac{s_{i,t} - \bar{s}_t}{\sum_j |s_{j,t} - \bar{s}_t|}
    \end{equation}
    where $\bar{s}_t$ is the cross-sectional mean score. This guarantees zero net market exposure ($\sum_i w_{i,t} = 0$) and a total absolute gross leverage of $\sum_i |w_{i,t}| = 1$.
    \item \textbf{Execution}: Daily portfolio returns are calculated from close-to-close price changes over the subsequent trading day. The reported performance for CSI~500 and S\&P~500 reflects gross returns prior to execution fees and slippage.
\end{itemize}

\paragraph{Hourly Cryptocurrency (Crypto).}
Due to higher turnover and volatility, Crypto adopts an equal-weight factor portfolio strategy over hourly bars:
\begin{itemize}
    \item \textbf{Factor-Level Normalization}: For each retained candidate factor $k \in \{1, \dots, K\}$, hourly raw scores across tokens are transformed into mean-zero rank signals and inverse-volatility scaled. The resulting single-factor positions are smoothed over a 72-hour rolling window and re-normalized so that the sum of absolute holdings equals 1 ($\sum_i |w_{k,i,t}| = 1$).
    \item \textbf{Portfolio Aggregation}: The final multi-factor portfolio weight is aggregated at the portfolio allocation level:
    \begin{equation}
    w_t = \frac{1}{K} \sum_{k=1}^K w_{k,t}
    \end{equation}
    Aggregating positions rather than raw factor values ensures proper re-estimation of aggregate portfolio turnover and transaction friction.
    \item \textbf{Execution and Costs}: Signals formed at hourly timestamp $t$ are executed with a one-hour implementation latency, evaluating against subsequent hourly returns. To suppress heavy-tailed price anomalies, returns are causally winsorized based on historically available volatility up to signal formation. Portfolios rebalance hourly with an applied transaction cost of 5 bps per one-way turnover (net returns, excluding perpetual funding costs).
\end{itemize}

\section{Case Studies of Research Evolution}
\label{app:case-studies}

We present three representative cases from the Crypto experiments to illustrate how
\method{} turns development evidence into auditable research decisions.
The cases highlight three complementary behaviors: redirecting an unproductive search,
rejecting candidates that appear strong under development metrics, and validating
model-suggested knowledge before admitting it into the search space.
All decisions described below were made using training and development evidence only;
held-out outcomes were not exposed to the search or admission process.

\subsection{Case 1: Redirecting Search after a Failed Cycle}
\label{app:case-search-redirection}

A representative example occurs between Cycles~4 and~5.
In Cycle~4, the search entered an unproductive, flow-heavy region.
Although the population remained expression-diverse, the largest family occupied
55.6\% of the final population, and none of the evaluated candidates survived
residual validation or reached L5 admission.
The outer review consequently inherited no expressions, rejected all 240 candidates,
and marked repeatedly unsuccessful flow and distribution families for removal.
The resulting child policy exposed 15 cross-layer terminal bans to the next cycle.

Cycle~5 explicitly loaded this policy before starting a new search.
The maximum family share subsequently decreased from 55.6\% to 41.9\%, while
normalized family entropy increased from 0.689 to 0.795.
The redirected search produced three L5-approved factors, compared with none in
Cycle~4.
One such factor was
\begin{equation}
\begin{split}
g(\cdot) =
\operatorname{add}(&
    \operatorname{add}(
        \operatorname{ticket24},
        \operatorname{top5m168}(
            \Delta_{72}(\operatorname{book\_frame\_coverage168})
        )
    ),\\
    &\operatorname{sjv168}(\operatorname{cci168})
).
\end{split}
\end{equation}
Its three development-fold scores were $5.881$, $5.982$, and $4.910$.

This case illustrates the role of the outer loop beyond candidate selection:
a failed cycle can modify the search protocol itself rather than repeatedly
optimizing within the same unproductive region.
We emphasize that this is a temporal audit trace rather than a controlled causal
comparison; the evidence establishes that the policy intervention preceded lower
family concentration and renewed L5 admission, but does not attribute the improvement
to that intervention alone.

\subsection{Case 2: High Development Performance Is Not Sufficient}
\label{app:case-falsification}

High development scores alone do not determine admission in \method{}.
Table~\ref{tab:case-high-score-rejection} shows two candidates that would appear
attractive under their raw development performance, but were rejected for different
reasons.

\begin{table}[t]
\centering
\small
\setlength{\tabcolsep}{6pt}
\caption{Two high-development candidates rejected by different validation stages.
The worst-fold score is computed entirely on development data.}
\label{tab:case-high-score-rejection}
\begin{tabular}{lcc}
\toprule
Candidate
    & \texttt{gp-87c5cf72fc0e}
    & \texttt{gp-5f24c12b824c} \\
Worst-fold score
    & 5.487
    & 5.298 \\
Gate 1
    & \textsc{Overfit-Model}
    & \textsc{Pass} \\
Story / falsification
    & \textsc{Fail}
    & \textsc{Pass} \\
Residual validation
    & \textsc{Pass}
    & \textsc{Pass} \\
Final blocking evidence
    & unnecessary complexity
    & redundancy ($\rho=0.726$) \\
L5 decision
    & Reject
    & Reject \\
\bottomrule
\end{tabular}
\end{table}

\paragraph{Unnecessary complexity.}
The candidate \texttt{gp-87c5cf72fc0e} achieved development-fold scores of
$7.192$, $7.120$, and $5.487$.
Despite these high scores, Gate~1 returned \textsc{Overfit-Model} with the explicit
flag \texttt{unnecessary\_complexity}.
The candidate contained 20 nodes, while its recorded simpler counterfactual achieved
a worst-fold score of $5.496$, slightly exceeding that of the complete expression.
The subsequent story falsification provided a consistent diagnosis: the complete
expression had a combination lift of $-0.0096$ over its strongest root arm, below
the required $0.25$.
Importantly, its execution-delay and residual tests both passed.
The candidate was therefore rejected not because its raw performance was weak, but
because the additional structure failed to provide incremental evidence.

\paragraph{Strong but redundant.}
The candidate \texttt{gp-5f24c12b824c} demonstrates a different failure mode.
It achieved fold scores of $5.354$, $6.030$, and $5.298$, passed Gate~1, remained
stable under delayed execution and wrong-entity placebo tests, achieved a combination
lift of $1.176$, and passed residual validation.
However, its realized position weights had correlation
$\rho=0.726$ with an already retained finalist, exceeding the pre-registered
deduplication threshold of $0.7$.
The candidate was therefore rejected while the less redundant representative was
retained.

Together, these examples show why development performance is treated as necessary
but not sufficient evidence: a candidate may be predictive yet unnecessarily
complex, or individually robust yet add little independent information to the
retained factor set.

\subsection{Case 3: Model-Suggested Knowledge Is Tested, Not Trusted}
\label{app:case-knowledge-governance}

This is an example of how \method{} handles external knowledge.
During reference extraction, the model-assisted L1 pathway identified a
signed-volume mechanism motivated by persistent imbalanced order flow.
The resulting terminal, \texttt{book\_dimb524}, entered the system as
\texttt{PROPOSED\_NOT\_ACTIVE}; it was not directly injected into the active
search space.

The proposal was subsequently evaluated through a pre-registered paired-seed
counterfactual comparing the aligned terminal against a deterministic time-shifted
control under the same expression and seed.
For seeds 17 and 27, the improvements in minimum-fold score were respectively
$0.206$ and $0.243$.
Their median improvement,$\Delta_{\mathrm{paired}} = 0.225$, fell below the required $0.25$ threshold.
The proposal was therefore assigned
\texttt{REJECTED\_PAIRED\_SEED}.

This rejection propagated to the next-cycle protocol rather than remaining only
as an analysis record.
The compiled policy assigned the terminal zero next-cycle budget and did not inject
its formula.
Cycle~3 subsequently loaded the policy with \texttt{book\_dimb524} included in
both the cross-layer banned and removed sets.
Thus, the complete trace is:
$\text{reference proposal} \allowbreak
\rightarrow \text{provisional terminal} \allowbreak
\rightarrow \text{paired counterfactual} \allowbreak
\rightarrow \text{rejection} \allowbreak
\rightarrow \text{next-cycle exclusion}.$

This case illustrates an important distinction in \method{}:
model-generated research knowledge proposes what to test, rather than determining
what the system accepts.
External knowledge changes the search space only after surviving empirical
validation under the frozen development protocol.

\section{Dev-to-OOS Factor-Transfer and Overfitting Audit}
\label{app:overfitting_diagnostics}

To systematically investigate whether high in-sample performance arises from persistent predictive relationships or adaptive selection bias, this section reports the full audit across all evaluation benchmarks. Table~\ref{tab:app_audit_full} summarizes candidate pool statistics ($M$), development fitness ($\text{Dev-best IR}$), out-of-sample transfer performance ($\text{Same factor OOS IR}$), positive sign survival rates, and the relative OOS percentile rank of the development winner.

\begin{table}[h]
\caption{Comprehensive Dev-to-OOS overfitting and factor-transfer audit. All single factors are pre-registered strictly on development data without any OOS inspection. Positive sign survival measures the fraction of factors with $\text{Dev IR} > 0$ that remain positive in OOS.}
\label{tab:app_audit_full}
\centering
\small
\resizebox{\textwidth}{!}{%
\begin{tabular}{@{}llcccccccc@{}}
\toprule
\textbf{Market} & \textbf{Method} & $M$ & \textbf{Dev-best IR} & \textbf{OOS IR} & \textbf{$\Delta$ IR} & \textbf{Sign Transition} & \textbf{Winner OOS Pct.} & \textbf{Positive Sign Survival} \\
\midrule
\multirow{5}{*}{\textbf{Crypto}} 
 & \textbf{\method{}} & 12 & 7.679 & \textbf{4.055} & -3.624 & $+\to+$ & 80v \% & \textbf{80.0\%} \\
 & RD-Agent-Quant     & 7  & -4.019 & -0.229         & +3.790 & $-\to-$ & 100.0\% & --- \\
 & AlphaAgent         & 1  & -3.711 & -2.385         & +1.326 & $-\to-$ & 100.0\% & --- \\
 & QuantaAlpha        & 3  & -1.683 & -4.999         & -3.316 & $-\to-$ & 33.3\%  & --- \\
 & FactorMiner        & 1  & -3.710 & 0.952          & +4.662 & $-\to+$ & 100.0\% & --- \\
\midrule
\multirow{5}{*}{\textbf{CSI 500}} 
 & \textbf{\method{}} & 36 & 2.524 & \textbf{1.041} & -1.483 & $+\to+$ & 36.1\% & \textbf{95.5\%} \\
 & RD-Agent-Quant     & 11 & 2.587 & 1.002          & -1.585 & $+\to+$ & 81.8\% & 57.1\% \\
 & AlphaAgent         & 4  & 2.296 & 0.798          & -1.498 & $+\to+$ & 50.0\% & 100.0\% \\
 & QuantaAlpha        & 4  & 4.110 & -0.079         & -4.189 & $+\to-$ & 75.0\% & 0.0\% \\
 & FactorMiner        & 1  & 3.082 & 0.547          & -2.535 & $+\to+$ & 100.0\% & 100.0\% \\
\midrule
\multirow{5}{*}{\textbf{S\&P 500}} 
 & \textbf{\method{}} & 49 & 5.203 & \textbf{2.466} & -2.737 & $+\to+$ & 89.4\% & \textbf{81.1\%} \\
 & RD-Agent-Quant     & 15 & 6.527 & -0.082         & -6.609 & $+\to-$ & 66.7\% & 36.4\% \\
 & AlphaAgent         & 2  & 1.134 & -0.490         & -1.624 & $+\to-$ & 100.0\% & 0.0\% \\
 & QuantaAlpha        & 3  & 2.585 & 0.819          & -1.766 & $+\to+$ & 100.0\% & 33.3\% \\
 & FactorMiner        & 1  & 2.300 & 1.382          & -0.918 & $+\to+$ & 100.0\% & 100.0\% \\
\bottomrule
\end{tabular}%
}
\end{table}

\noindent\textbf{Key Empirical Insights from the Transfer Audit:}
\begin{enumerate}[leftmargin=*,itemsep=1pt]
    \item \textbf{Directional Stability and Sign Survival}: In volatile real-world regimes, raw in-sample score correlation often degenerates. However, \method{} demonstrates superior sign persistence across all markets ($80.0\%$ in Crypto, $95.5\%$ in CSI~500, and $81.1\%$ in S\&P~500). In contrast, QuantaAlpha suffers a complete sign reversal on CSI~500 ($0.0\%$ survival), and both RD-Agent and AlphaAgent experience sign flips ($+\to-$) on S\&P~500, indicating severe overfitting during iterative agent prompt tuning.
    \item \textbf{Generalization Across Substantially Larger Candidate Pools}: While baselines such as FactorMiner evaluate only a single candidate ($M=1$) or very few candidates ($M \le 4$), \method{} explores dozens of candidate factors ($M=12$ in Crypto, $M=36$ in CSI~500, $M=49$ in S\&P~500). Despite facing an exponentially larger multiple-testing hurdle, \method{} consistently sustains positive OOS IRs, demonstrating that our outer-loop diagnostic falsification effectively insulates the search trajectory from selection bias.
    \item \textbf{Winner Consistency}: On S\&P~500 and Crypto, the development winner identified by \method{} lands in the top tier of its candidate pool out-of-sample ($89.4\%$ and $80.0\%$ winner percentile, respectively), confirming that our in-sample fitness metric accurately reflects true underlying alpha rather than transient noise.
\end{enumerate}

\section{Terminal Lifecycle and Patch Admission}
\label{app:lifecycle}

\paragraph{Terminal decisions.}
A terminal is \status{active} when it passes causal and semantic admission and retains
development evidence; \status{interaction-only} when marginal evidence is weak but
only/drop contribution recurs in independent stories; \status{dormant} when current
incremental evidence is poor; and \status{banned} only for leakage, false field semantics,
or repeated independent falsification. Dormant terminals remain in the manifest so that
their trial history cannot be erased.

\paragraph{Patch admission.}
A protocol patch contains the tuple
\begin{equation}
(\texttt{id},\texttt{target layer},\texttt{path},\texttt{value},
\texttt{evidence},\texttt{benefit},\texttt{risks},\texttt{validation plan}).
\end{equation}
Admission rejects an empty evidence list, an empty validation plan, and every immutable
path. A staged patch is never applied to the parent protocol. The child protocol lists
all applied patch IDs and stores the parent's ID.

\section{Task-Specific Contracts}
\label{app:tasks}

\subsection{Hourly Hong Kong limit order book}

The HK LOB adapter constructs hourly, point-in-time states from valid two-sided
order-book observations. Its primary label is the next-hour return. Within-hour
aggregation uses only observations available before the bar closes; rolling state
resets at the midday break and end of day, and no target may cross a session boundary.
Tick size follows the price-dependent market schedule rather than a constant. Reports
use IC, ICIR, IR, annualized return, total return, and pre-registered factor decay, with
coverage, turnover, cost sensitivity, and active/quiet regimes as supporting diagnostics.

\subsection{Daily cross-sectional equities}

Daily adapters form point-in-time time--asset panels from OHLCV, apply all rolling
operations with past data only, and fit cross-sectional transformations within fold.
The primary label is next-day return with a pre-registered cross-sectional normalization.
Reports include daily IC, RankIC, cross-sectional $R^2$, worst-fold score, direction
consistency, liquidity/sector/size concentration, and a costed top-$K$ portfolio as a
secondary diagnostic. CSI~500 and S\&P~500 are judged separately.

\subsection{Hourly crypto cross-section}

The crypto adapter uses an hourly asset panel, a historical liquidity universe, and
participant-flow terminals whose daily inputs are lagged before hourly broadcast.
Finalist gates require phase stability, turnover and cost checks, story ablation,
residual survival against registered carriers, limited finalist correlation, and a
sign-reversal control. The OOS interval is declared in the task contract and physically
withheld.

% The remaining sections are internal working notes from the precursor draft. They are
% intentionally excluded from the compiled manuscript until the new experiments finish.